\documentclass{article}
\PassOptionsToPackage{numbers, compress}{natbib}
\makeatletter
\providecommand{\@runningtitle}{The Era of Agentic Organization: Learning to Organize with Language Models}
\makeatother
\usepackage[preprint]{neurips_2025_custom}
\usepackage{fix-cm}

\usepackage{xcolor}         
\definecolor{linkColor}{rgb}{0.2,0.4,0.6}
\definecolor{myblue}{HTML}{0379AC}
\definecolor{myred}{HTML}{A50E50}
\usepackage[utf8]{inputenc} 
\usepackage[T1]{fontenc}    
\usepackage[colorlinks=true,linkcolor=linkColor,citecolor=linkColor,filecolor=linkColor,urlcolor=linkColor]{hyperref}       
\usepackage{url}            
\usepackage{booktabs}       
\usepackage[table]{xcolor}
\usepackage{amsfonts}       
\usepackage{nicefrac}       
\usepackage{microtype}      
\usepackage{natbib} 
\usepackage{graphicx}
\usepackage{arydshln}
\usepackage{booktabs}
\usepackage{multirow}
\usepackage{caption}
\usepackage{subcaption}
\usepackage{makecell}
\usepackage{csquotes}
\usepackage{epigraph}

\usepackage{amsthm}
\tcbuselibrary{listings, breakable, skins}
\RequirePackage{algorithm}
\RequirePackage{algorithmic}

\usepackage{multirow}
\usepackage{amsmath}
\usepackage{capt-of}
\usepackage{tabularx}
\usepackage{epsfig}
\usepackage{amssymb}
\usepackage{amsfonts}
\usepackage{booktabs}
\usepackage{scalerel}
\usepackage[inline]{enumitem}
\usepackage{listings}
\usepackage{varwidth}
\usepackage[export]{adjustbox}
\usepackage{tikz}
\usetikzlibrary{tikzmark}

\usepackage{stmaryrd}
\usepackage{bbm}
\usepackage{wrapfig}
\usepackage{pifont}
\usepackage[noabbrev]{cleveref}

\definecolor{deepblue}{rgb}{0,0,0.5}
\definecolor{officeblue}{RGB}{0,102,204}
\definecolor{deepred}{rgb}{0.6,0,0}
\definecolor{deepgreen}{rgb}{0,0.5,0}
\definecolor{mybrickred}{RGB}{182,50,28}

\definecolor{fillcolor}{RGB}{216,217,252}

\usepackage{etoolbox}
\usepackage{framed}

\newif\ifxetexorluatex
\ifxetex
  \xetexorluatextrue
\else
  \ifluatex
    \xetexorluatextrue
  \else
    \xetexorluatexfalse
  \fi
\fi
\newcommand*\quotesize{60} 
\newcommand*{\openquote}
   {\tikz[remember picture,overlay,xshift=-4ex,yshift=-2.5ex]
   \node (OQ) {\fontsize{\quotesize}{\quotesize}\selectfont``};\kern0pt}

\newcommand*{\closequote}[1]
  {\tikz[remember picture,overlay,xshift=4ex,yshift={#1}]
   \node (CQ) {\fontsize{\quotesize}{\quotesize}\selectfont''};}

\colorlet{shadecolor}{white}

\newcommand*\shadedauthorformat{\emph} 

\newcommand*\authoralign[1]{%
  \if#1l
    \def\authorfill{}\def\quotefill{\hfill}
  \else
    \if#1r
      \def\authorfill{\hfill}\def\quotefill{}
    \else
      \if#1c
        \gdef\authorfill{\hfill}\def\quotefill{\hfill}
      \else\typeout{Invalid option}
      \fi
    \fi
  \fi}
{\authoralign{#1}
\ifblank{#2}
   {\def\shadequoteauthor{}\def\yshift{-2ex}\def\quotefill{\hfill}}
   {\def\shadequoteauthor{\par\authorfill\shadedauthorformat{#2}}\def\yshift{2ex}}
\begin{snugshade}\begin{quote}\openquote}
{\shadequoteauthor\quotefill\closequote{\yshift}\end{quote}\end{snugshade}}

\newfloat{Code}{htbp}{Code}

\usepackage{amsmath,amsfonts,bm}

\def\eqref#1{equation~\ref{#1}}

\def\1{\bm{1}}

\DeclareMathAlphabet{\mathsfit}{\encodingdefault}{\sfdefault}{m}{sl}
\SetMathAlphabet{\mathsfit}{bold}{\encodingdefault}{\sfdefault}{bx}{n}

\theoremstyle{plain}

\theoremstyle{definition}

\theoremstyle{remark}

\newcommand\our{AsyncThink}
\newcommand\org{organizer}
\newcommand\worker{worker}
\definecolor{verylightgray}{rgb}{0.93, 0.93, 0.93}

\title{The Era of Agentic Organization: \\ Learning to Organize with Language Models}

\author{
Zewen Chi~~~Li Dong
~\bf Qingxiu Dong~~~Yaru Hao~~~Xun Wu~~~Shaohan Huang~~~Furu Wei\thanks{Contact person: \href{mailto:fuwei@microsoft.com}{fuwei@microsoft.com}.} \\
Microsoft Research \\
~{\href{https://aka.ms/GeneralAI}{https://aka.ms/GeneralAI}}
}

\begin{document}

\maketitle

\begin{abstract}
We envision a new era of AI, termed \textbf{agentic organization}, 
where agents solve complex problems by working collaboratively and concurrently, enabling outcomes beyond individual intelligence.
To realize this vision, we introduce \textbf{asynchronous thinking} (\our{}) as a new paradigm of reasoning with large language models, which organizes the internal thinking process into concurrently executable structures. Specifically, we propose a thinking protocol where an organizer dynamically assigns sub-queries to workers, merges intermediate knowledge, and produces coherent solutions. More importantly, the thinking structure in this protocol can be further optimized through reinforcement learning. Experiments demonstrate that \our{} achieves 28\% lower inference latency compared to parallel thinking while improving accuracy on mathematical reasoning. Moreover, \our{} generalizes its learned asynchronous thinking capabilities, effectively tackling unseen tasks without additional training.
\end{abstract}

\begin{figure*}[h!]
\centering
\includegraphics[width=0.85\linewidth]{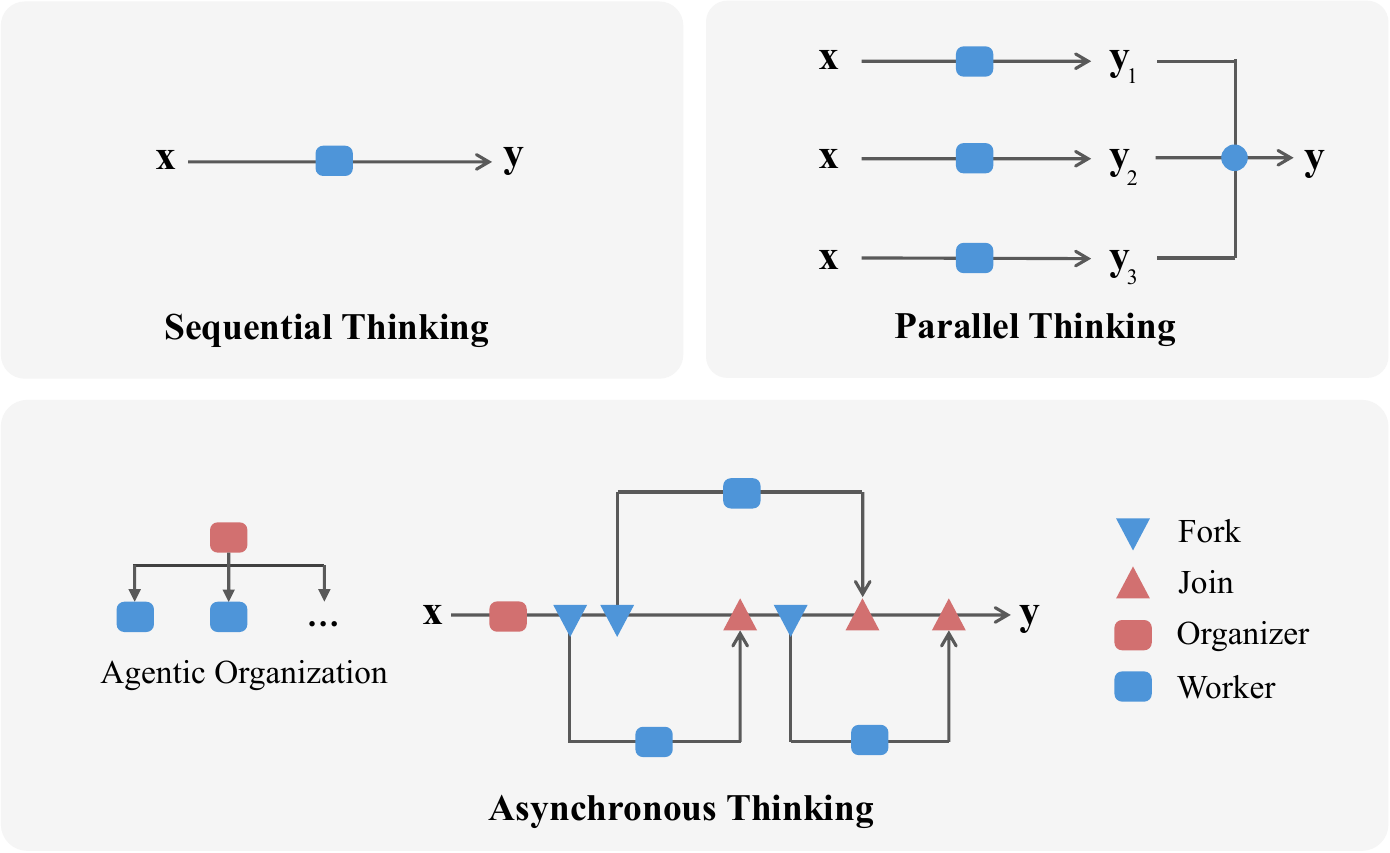}
\caption{Comparison among our asynchronous thinking paradigm, sequential thinking, and parallel thinking. Sequential thinking employs a purely sequential decoding trajectory; parallel thinking executes multiple independent traces with an outcome aggregation. Differently, \our{} learns to form an agentic organization to think concurrently and collaboratively.}
\label{fig:fig1}
\end{figure*}

\section{Introduction}
\label{sec:intro}

Our vision for the next era of artificial intelligence is to realize \textbf{agentic organization}, where agents form organizational systems that collaborate to tackle complex problems beyond the limits of individual intelligence~\cite{Casadei2023ArtificialCI,Pretel2025DigitalTA}.
Although large language models have unlocked remarkable reasoning capabilities as individual agents~\cite{ElKishky2024OpenAIO1, Bai2025KimiK2}, realizing this vision requires reasoning models that can not only think independently but also think collaboratively as an organized system. This gap motivates research into new reasoning paradigms~\cite{parallel-r1,apr}.

Developing AI systems that realize agentic organization introduces several key challenges~\cite{Han2024LLMMS}. First, organizing multiple agents to think concurrently often incurs additional latency~\cite{Zhang2024CutTC}. Current parallel thinking approaches, which typically generate independent thinking traces and aggregate them afterward~\cite{yao2023tree,zhao2025majority,brown2024large}, are limited not only by the slowest thinking trace but also by the additional delay incurred during the final aggregation process. Second, adaptivity and dynamism remain difficult to achieve. Parallel thinking methods rely on manually designed, fixed workflows that cannot accommodate the diverse requirements of different queries~\cite{apr}. Some tasks benefit from divide-and-conquer strategies, while others require step-by-step reasoning. Third, learning effective agentic organization policies remains an open problem~\cite{orga}, as manually designing optimal thinking structures for every possible query is intractable. 

In this work, we introduce \textbf{asynchronous thinking} (\our{}) as a new paradigm for reasoning with large language models, with the goal of \textbf{learning to organize} the internal thinking into concurrently executable structures. Specifically, we propose a thinking protocol where an LLM plays both roles: an organizer that dynamically structures the process through \texttt{Fork} and \texttt{Join} actions, and workers that execute sub-queries and return intermediate knowledge or results. This thinking protocol provides the foundation for adaptivity and dynamic reasoning, allowing the model to explore diverse execution structures.

To train an \our{} model, we propose a two-stage training procedure. First, we perform a cold-start format fine-tuning on synthetic role-specific data to learn the syntax of \our{} actions. Then, we further train the \our{} model with a reinforcement learning stage, with rewards that encourage correctness, format compliance, and thinking concurrency.

We evaluate our \our{} models on multi-solution countdown, mathematical reasoning, and Sudoku tasks. Our experiments demonstrate that our models consistently achieve higher accuracy while reducing latency compared to sequential thinking and parallel thinking models. We further analyze its performance through ablation studies and accuracy-latency frontier comparisons, highlighting the effectiveness of our two-stage training procedure. Remarkably, our models also demonstrate strong generalization, exhibiting zero-shot \our{} reasoning on previously unseen tasks despite training solely on simple countdown data.

Our contributions are as follows:
\begin{itemize}[leftmargin=*]
\setlength\itemsep{0.01em}
\item We formalize the \textbf{learning-to-organize} problem, clarifying its setup and objective, building a foundation for studying how to organize agents to collaborate and operate concurrently.
\item  We introduce \textbf{asynchronous thinking}, a new reasoning paradigm that allows large language models to organize their internal thinking into concurrently executable structures, and learn this organization ability through reinforcement learning.
\item We evaluate \our{} models on reasoning tasks, demonstrating improved accuracy and reduced latency. Remarkably, our models generalize asynchronous thinking to unseen tasks.
\end{itemize}

\begin{figure*}[t]
    \centering
    \includegraphics[width=1.0\linewidth]{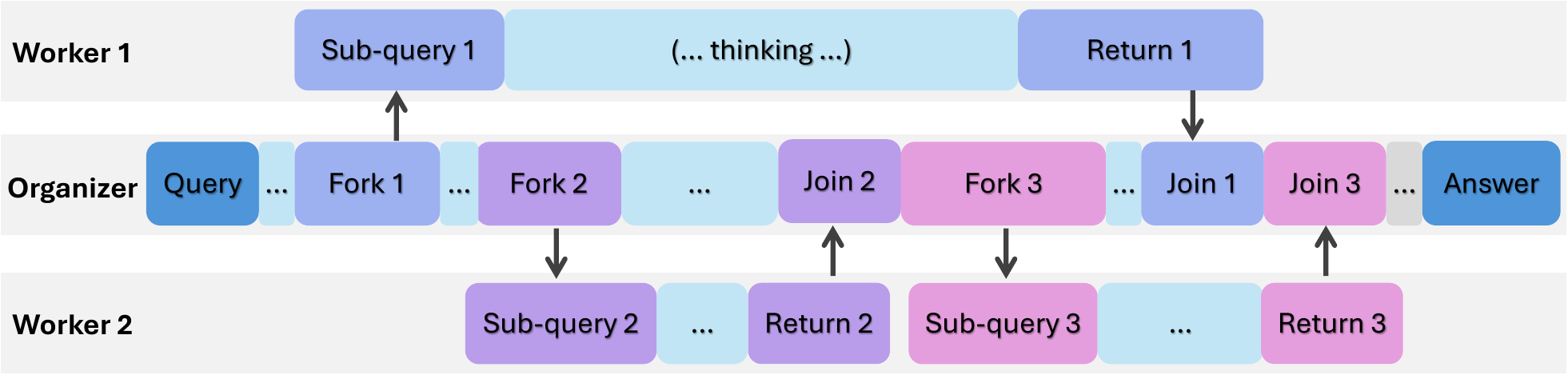}
    \caption{An example of the thinking protocol of \our{}. The protocol achieves asynchronous thinking through the \texttt{Fork}-\texttt{Join} actions, which controls the trajectory of thinking.}
    \label{fig:method-main}
\end{figure*}

\begin{table}[t]
\centering
\rowcolors{2}{gray!15}{white}
\scalebox{1.0}{
\begin{tabular}{p{0.18\linewidth} p{0.35\linewidth} p{0.35\linewidth}}
\toprule
\textbf{Concept} & \textbf{Definition} & \textbf{Analogue (Computer Systems)} \\
\midrule

\emph{Agent}
& A model performing actions sequentially.
& \emph{CPU core} that executes instructions for a single process at a time. \\

\emph{Agent Pool}
& A group of agents, determining the number of agents that can run concurrently.
& \emph{Multicore CPU} representing all available cores and defining parallel capacity. \\

\emph{Organization \newline Policy}
& The strategy of organizing the agents to work collaboratively and concurrently.
& \emph{Multiprocess program} that organizes processes to achieve optimized execution. \\

\bottomrule
\end{tabular}
}
\caption{Analogy between agentic-organization concepts and computer system components.}
\label{tab:analogy}
\end{table}

\section{Organizer-Worker Thinking Protocol}
\label{sec:inference_engine}

Figure~\ref{fig:method-main} illustrates an overview of our asynchronous thinking paradigm.
The organizer-worker thinking protocol introduces two classes of roles for thinking, where an organizer manages the thinking processes and workers execute individual thinking processes for the assigned sub-queries.
The two agent roles, \org{} and \worker{}, share the same LLM backbone and both perform autoregressive text decoding. Their distinction lies primarily in the set of actions each can take. This protocol provides the foundation for \our{} to explore diverse thinking structures.

\subsection{Agentic Organization}

Table~\ref{tab:analogy} outlines the concepts of agentic organization by drawing an analogy to those of computer systems. Specifically, an \textit{agent} refers to a model that can perform actions sequentially, and an \textit{agent pool} represents a group of agents that run concurrently. In our setup, we specify an agent pool with a fixed capacity to ensure fair comparisons between different methods. In addition, agents may share the same model and model weights, as they can be viewed as different instances of the same model. \textit{Organization policy} refers to the strategy of organizing agents to work collaboratively and concurrently to complete tasks. For example, a straightforward design has agents that operate independently and aggregate their outcomes in the end, as in parallel thinking \cite{zhao2025majority,wen2025parathinker}. In contrast, in this work, we propose an adaptive organization policy rather than a fixed, manually designed one.

\subsection{Organizer}

Organizer is the main role running through the whole thinking process, which is responsible for global coordination of asynchronous thinking. At the beginning of thinking, we concatenate the system prompt, which instructs the LLM how to take the \texttt{Fork}-\texttt{Join} actions in the text format, and the user query as the prompt. During the process, an \org{} can take one of the following four actions sequentially:
\begin{itemize}[leftmargin=*]
\setlength\itemsep{0.01em}
\item \texttt{Think}: Advance the current decoding process in its own decoding thread.
\item \texttt{Fork}: Assign a thinking job to an available \worker{} with a sub-query. The expected format is 
\colorbox{verylightgray}{
... \textlangle FORK-$i$\textrangle~\textit{sub-query} \textlangle /FORK-$i$\textrangle~ ...
}, where $i$ is a sub-query identifier used to distinguish this sub-query from the others. When a sub-query $i$ already exists or there are already $c - 1$ active \worker{}s, \org{} cannot take the \texttt{Fork}-$i$ action.
\item \texttt{Join}: Request the output of a previously \texttt{Fork}-ed thinking job, with the format of \colorbox{verylightgray}{ ... \textlangle JOIN-i\textrangle}.
Upon recognizing this request, the generation process of the \org{} may pause and wait for the corresponding \worker{} to finish thinking on the assigned sub-query $i$. The returned output from the \worker{} is then appended to the tail of the current \org{} decoding context.
The context format after completion is
\colorbox{verylightgray}{
... \textlangle JOIN-i\textrangle~\textit{returned text here} \textlangle /JOIN-i\textrangle~ ...
}.
\item \texttt{Answer}: Terminate the inference process and produce the final answer.
\end{itemize}

The above actions can be generated in a pure text format, making them directly compatible with the input-output formats of existing LLMs.

\subsection{Worker}
Within an agent pool with capacity of $c$, there are  $c-1$ \worker{}s, which receive requests from the \org{} and independently carry out thinking tasks for answering sub-queries.
Once a \worker{} completes the thinking for a sub-query and \org{} requests the corresponding result, it sends the result back to the \org{}, enabling communication.
We consider a \worker{} active when it is engaged in processing a sub-query and has not yet returned its result. 
Similar to the \org{}, the input of a worker is a combination of the system prompt and the sub-query received from the \org{}. The expected output format is 
\colorbox{verylightgray}{
... \textit{thoughts} ... \textlangle RETURN\textrangle~\textit{some takeaways}\textlangle RETURN\textrangle~...
}.


\subsection{Asynchronous Thinking}

Under the organizer-worker thinking protocol, the thinking procedure of \our{} begins with an \org{}. This is the main role that receives the user query and drives the entire thinking process.
Specifically, the \org{} generates text tokens and action tags autoregressively, without using explicit action-selection modules.

When a \texttt{Fork} tag appears, \our{} proposes a sub-query and assigns it to an available \worker{}. Upon receiving the request, the \worker{} begins thinking for the sub-query. At the same time, the \org{} continues its own execution such as thinking for \texttt{Fork}-ing another job, while \worker{}s run concurrently. Upon encountering a \texttt{Join} tag, \our{} synchronizes with the corresponding \worker{}, i.e., if the \worker{} is still running, the \org{} pauses its decoding process until the \worker{} returns results. Otherwise, the returned text will be appended to the \org{} LLM decoding context, and the \org{} resumes executing.

The thinking process concludes when the \org{} generates a final answer ending with an end-of-sentence token. The thinking protocol operates entirely at the input-output surface of LLMs and does not require any modification to the underlying neural network architectures. Notably, it allows open-ended generation of the execution structure of thinking, providing a unified formulation under which existing thinking paradigms emerge as special cases. For example, sequential thinking can be realized when no \texttt{Fork} actions are taken, while parallel thinking arises when the \org{} repeatedly assigns the original query to multiple \worker{}s.

\section{Learning to Organize}

We train \our{} through a two-stage procedure. In cold-start format fine-tuning, we synthesize training data for the thinking protocol, curating query-response pairs that teach the model to play both roles. Subsequently, in reinforcement learning, the model explores diverse thinking structures and refines its asynchronous thinking capabilities, guided by reward signals that evaluate both reasoning quality and efficiency.

\subsection{Stage 1: Cold-Start Format Fine-Tuning}
\label{sec:data_construction}

This section introduces how to start training an \our{} model from an existing LLM with the cold-start format fine-tuning stage.

\paragraph{Data Synthesis} Since existing corpora rarely contain organizer-worker thinking traces, we synthesize training data with the GPT-4o model. We use GPT-4o to analyze each query to detect conditionally independent thinking fragments. The model then generates the traces of both roles, following the organizer-worker thinking protocol formats by providing few-shot examples. Then, we verify the synthesized data with rules and filter out the data with format errors.

\paragraph{Random Initialization of Thinking Structure}
We empirically find that within an agent pool of $c > 2$, the GPT-4o model predominantly produces the \org{} traces that follow two distinct thinking topologies: (1) interleaved \texttt{Fork} and \texttt{Join} operations, resulting in only one active \worker{} at a time, and (2) a sequential pattern that first performs \texttt{Fork} $c - 1$ times, followed by \texttt{Join} $c - 1$ times. However, relying on a single or two thinking topologies reduces the plasticity of the model during training, and limits its ability to explore diverse organization policies. To enable broader exploration, we randomly sample \org{} action sequences and add one of the sequences to the prompt to guide the model to generate \org{} outputs following the sampled structure.

\paragraph{Supervised Format Fine-Tuning}
We split the synthesized training data into query-response pairs and then fine-tune the LLM using the standard causal language modeling objective. This supervised format fine-tuning phase equips the model with the ability to emit valid \org{} actions. At this stage, the model has not yet learned to produce correct answers with asynchronous thinking but only mimics the format. This limitation motivates the subsequent reinforcement learning stage.

\begin{figure*}[t]
    \centering
    \includegraphics[width=0.8\linewidth]{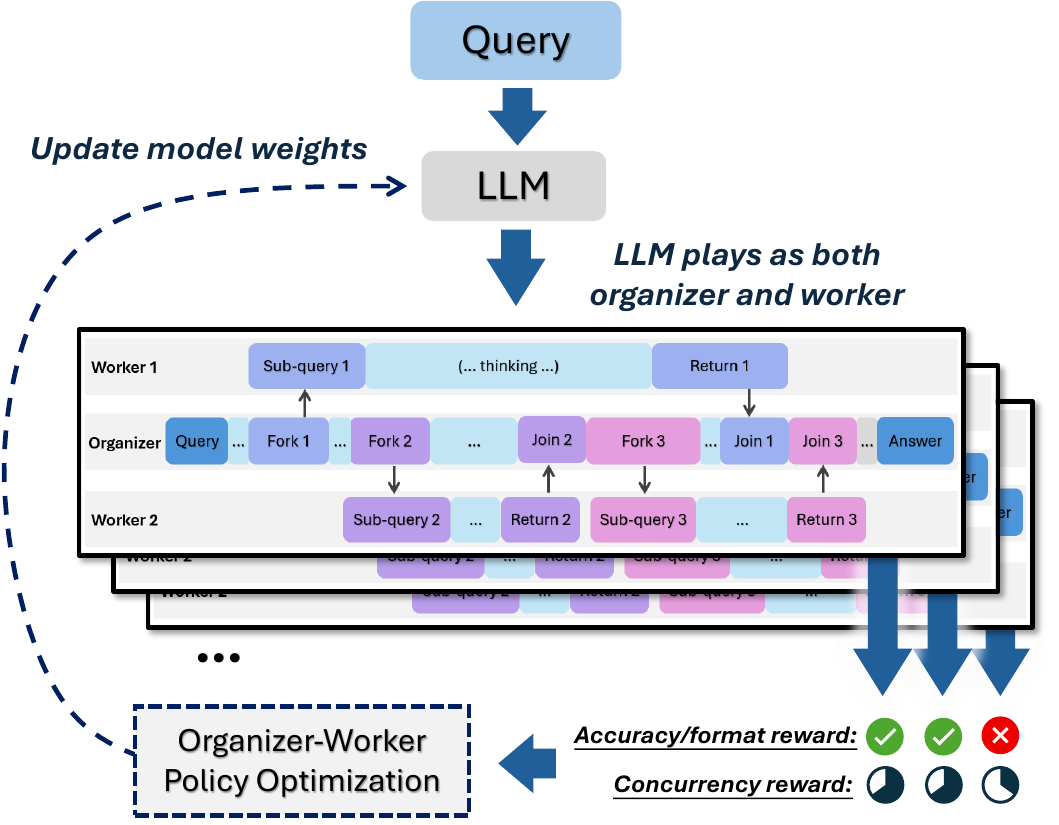}
    \caption{Illustration of the reinforcement learning framework for \our{}.}
    \label{fig:rl}
\end{figure*}

\subsection{Stage 2: Reinforcement Learning}

Since the cold-start format fine-tuning stage only teaches the syntactic structure of \org{} actions, the model still lacks the capability to exploit such thinking mechanism to produce final answers. Thus, we further optimize the model using reinforcement learning to improve asynchronous thinking.

\paragraph{Reward Modeling} 
We utilize a rule-based reward system that provides the following three types of rewards, encouraging both final-answer accuracy and thinking efficiency.
\begin{itemize}[leftmargin=*]
\setlength\itemsep{0.01em}
\item \textbf{Accuracy Reward}: This reward measures the accuracy of the predicted final answers, typically a binary reward for single-answer questions with $R_\text{A}=1$ for success and $R_\text{A}=0$ for failure. For multi-solution questions, we use $R_\text{A}=\min(n_c, n_s)/n_s$ as the reward, where $n_c$ and $n_s$ represent the number of unique and correct answers and the total required solutions, respectively.
\item \textbf{Format Reward}: We penalize the \org{} outputs that have format errors, directly assigning a constant format error reward $R_\text{FE}$ if the \org{} produces the following format errors, including (1) duplicated sub-query index when the \org{} redundantly \texttt{Fork}s a sub-query $i$ when another sub-query $i$ already exists and being processed, (2) agent pool overflow when the \org{} \texttt{Fork}s a new thinking job and there are already $c-1$ active \worker{}s, (3) \texttt{Join}-ing a non-existing sub-query, and (4) stopping without generating a final answer.
\item \textbf{Thinking Concurrency Reward}: This reward encourages the model to efficiently organize the thinking processes into concurrently executable parts. Specifically, let $a_t$ denote the number of active workers in an agent pool with capacity of $c$ at global decoding step $t$, where a worker is considered active if it has been delegated a sub-query and is currently engaged in LLM decoding. We define thinking concurrency ratio as
\begin{align}
    \eta = \frac{1}{T} \sum_{t=1}^{T} a_t,
\end{align}
where $T$ is the critical-path latency that will be introduced in Section~\ref{sec:compute_latency}. Next, we compute the thinking concurrency reward as
\begin{align}
    R_\eta=\min(\eta/c,\tau) / \tau,
\end{align}
where $\tau$ is a configurable threshold that prevents the model from hacking thinking concurrency, which is usually easier than predicting correct answers. 
The overall reward for an \our{} thinking trace $i$ is computed as
\begin{align}
R_i=
\begin{cases}
R_\text{FE} &, \text{trace $i$ has format errors} \\
R_\text{A} + \lambda R_\eta &, \text{otherwise}
\end{cases}
\end{align}
\end{itemize}

\paragraph{Organizer-Worker Policy Optimization}
We extend the group relative policy optimization (GRPO) \cite{grpo} to handle the non-sequential thought samples in the reinforcement learning for \our{}.
In common outcome-supervised setups, the episode is a single sequence of tokens. Differently, an episode of \our{} comprises multiple output traces from the \org{} and its associated \worker{}s. To accommodate this structure, we treat the \org{} trace together with its corresponding \worker{} traces as a single unit when computing rewards and group-relative advantages. The resulting shared advantage is then assigned to all tokens produced by both roles. Note that tokens in the \org{} trace that correspond to \worker{}-returned outputs, i.e., segments formatted as `\colorbox{verylightgray}{\textit{returned text here} \textlangle /JOIN-i\textrangle}') are masked out when computing the loss, as they are not produced by the \org{} itself. The initial \texttt{Join} tag tokens, which merge the returned messages into the \org{} context, are included in the loss computing for \org{} traces.

\begin{figure*}[t]
    \centering
    \includegraphics[width=1.0\linewidth]{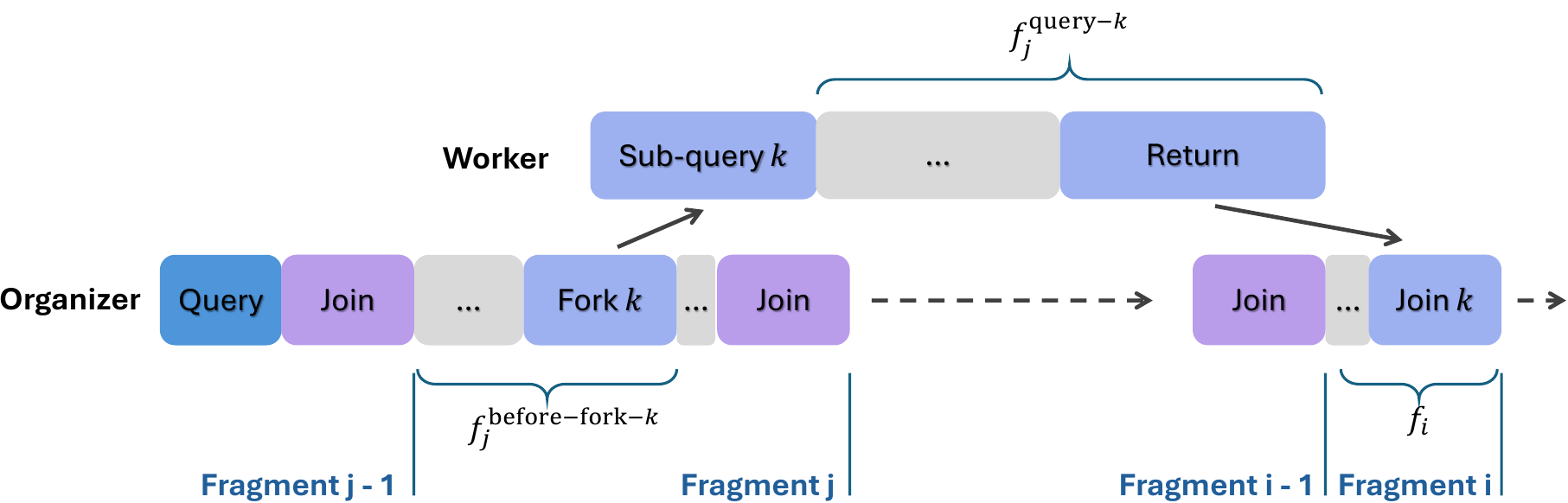}
    \caption{Illustration of the subproblem structure used in the dynamic programming formulation for computing the critical-path latency of an \our{} thinking trajectory.}
    \label{fig:critical-path}
\end{figure*}

\section{Experiments}

To evaluate \our{}, we conduct experiments on three tasks, including multi-solution countdown, math reasoning, and Sudoku.
In this section, we first introduce the evaluation metrics for asynchronous thinking, and then present the experimental details for each task individually.

\subsection{Evaluation Metrics}
Different from the evaluation of sequential reasoning, asynchronous thinking evaluation requires considering both overall correctness and thinking efficiency. 
We use the following two metrics for asynchronous thinking: (1) \textit{Final-Answer Accuracy} measures the correctness of the final answers. This metric reflects the overall reasoning capability of the model and enables direct comparison with other LLM reasoning methods.
(2) \textit{Critical-Path Latency}\label{sec:compute_latency} measures the minimum sequential depth required for asynchronous thinking. It represents a theoretical lower bound on inference time, abstracting away from implementation details such as the underlying inference engine implementation or hardware differences.

Figure~\ref{fig:critical-path} shows how to compute the critical-path latency of an \our{} thinking trajectory. According to the organizer-worker thinking protocol, the thinking trajectory takes the form of as a directed acyclic graph, where the \org{} might wait for \worker{}s to finish at the \texttt{Join} tags, introducing extra latency beyond the latency of the \org{} generating its own outputs. We present a dynamic programming method to compute the overall \our{} inference latency by breaking it down into a simpler subproblem, i.e., computing the latency at each \texttt{Join} position. Specifically, we split the \org{} output sequence at the \texttt{Join} positions, and obtain $n_J + 1$ fragments, where $n_J$ is the number of \texttt{Join} tags. The latency of finishing the $i$-th fragment $l_i$ is computed as
\begin{align}
l_i = \begin{cases}
    0 &, i=0 \\
    \max(l_{i-1}+f_i, l_{j-1} + f_j^{\text{before-fork-}k} + f^{\text{query-}k}) &, 1 \leq i \leq n_J \\
    ~l_{n_J} + f_{n_J + 1} &, i = n_J + 1
\end{cases}
\end{align}
where we denote $f_i$ the number of decoding steps of the $i$-th fragment; $j$ stands for the fragment index of the \texttt{Fork} that the current \texttt{Join} is associated with; $k$ represents the sub-query index of that \texttt{Fork}; $f_j^{\text{before-fork-}k}$ is the number of decoding steps within the fragment $j$ to finish the \texttt{Fork} tags; $f^{\text{query-}k}$ stands for the number of decoding steps to response to the sub-query $k$.

\subsection{Multi-Solution Countdown}

The countdown task is a widely used testbed for evaluating the reasoning capabilities of LLMs, with the goal of finding the arithmetic operations ($+, -, \times, /$) that convert the given numbers into the target number. Since the typical countdown task is not challenging for current LLMs, we extend the task to a harder version, named \textbf{multi-solution countdown} (MCD). Given a set of numbers, the goal is to find exactly four different solutions using three to six numbers from a given set of numbers. Here we define that two solutions are different when they have at least one different number or have at least one arithmetic operation used different times.

\paragraph{Data}
To construct the dataset, we randomly select $100$ unique numbers ranging from $1$ to $1,000$ as the target numbers of the test set. For each target number, we sample $4$ unique subsets of $\{1, ..., 100\}$, which leads to $400$ test examples. We ensure that the subsets have at least $4$ different solutions. Similarly, we also construct training data with $22,500$ examples whose target numbers are different from those in the test set. In addition, we synthesize the cold-start data with an agent pool capacity of $c=2$, following the procedure described in Section~\ref{sec:data_construction}. The resulting training corpus contains $25.5$K query-response pairs. Notice that an \our{} inference trace produces $2n_q+1$ query-response pairs, where $n_q$ is the number of sub-queries proposed by the \org{}.

\paragraph{Training}
We first perform the supervised format fine-tuning (SFT) on the Qwen3-4B \cite{qwen3} base model for $1$K steps, with a learning rate of $10^{-6}$ and a batch size of $128$. Then, we train the SFT-ed model with reinforcement learning (RL) following the organizer-worker thinking protocol to sample asynchronous thinking trajectories with an agent pool capacity of $c=2$. During RL, we use a learning rate of $10^{-6}$ and a batch of $64$ MCD examples, and sample $32$ trajectories for each query to compute group-relative advantages. We use the ratio of valid solutions found by the models as the accuracy reward for the MCD task. We set the maximum response length for \our{} \worker{}s to $2$K tokens for answering each sub-query. The maximum response length is limited to $2$K tokens prior to each \org{} \texttt{Join} action, after which the decoding budget resets to another $2$K tokens.

\paragraph{Baselines}
We post-train the Qwen3-4B \cite{qwen3} base model for sequential thinking using reinforcement learning, where the model is trained to answer the query after thinking without \texttt{Fork}s or \texttt{Join}s. In addition, we also include a parallel thinking baseline that samples two responses from the sequential-thinking models with the same query separately. Then, it performs majority vote to select the first four most frequent solutions as the final answer, from around eight predicted solutions.

\begin{figure*}[t]
    \centering
    \includegraphics[width=0.85\linewidth]{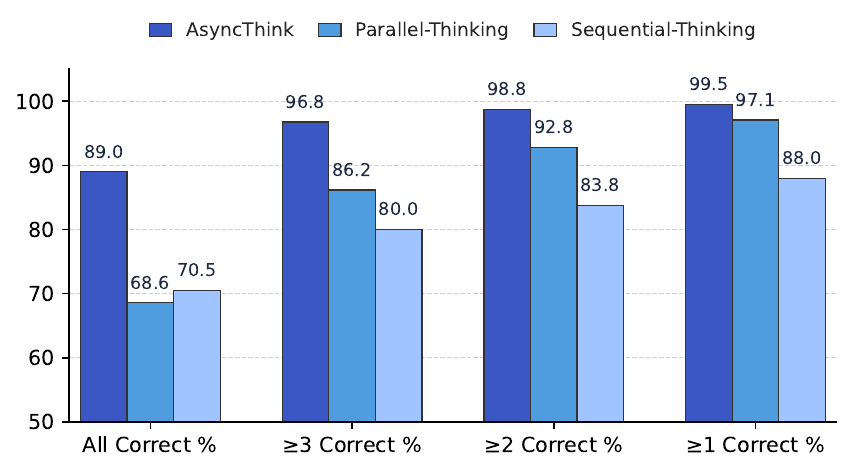}
    \caption{Evaluation results on the multi-solution countdown task. `$\geq a$ \texttt{Correct}' measures whether the model successfully finds $a$ unique and correct solutions for a given question. Results are averaged across $5$ random seeds.}
    \label{fig:mcd_main_bar}
\end{figure*}

\paragraph{Results}
Figure~\ref{fig:mcd_main_bar} presents the accuracy scores of the evaluated methods on the multi-solution countdown task, where `$\geq a$ \texttt{Correct}' means that the answer is considered correct if the model successfully finds $a$ unique and correct solutions for a given question. The figure shows that \our{} substantially outperforms the baseline methods in all correctness thresholds. In particular, when evaluating strict correctness, i.e., `\texttt{All Correct}', \our{} achieves $89.0\%$ against $68.6\%$ and $70.5\%$ of the baseline methods. The improvement demonstrates that our asynchronous thinking method not only enhances the overall accuracy of the predicted countdown solutions but also leads to more reliable multi-solution coverage.

\subsection{Math Reasoning}
We also assess \our{} on math reasoning benchmarks, including AMC-23 and AIME-24~\cite{aime2024}.

\paragraph{Data} We use the DeepScaleR dataset \cite{deepscaler} as the RL training data for both \our{} and the baseline methods. We synthesize the \our{} cold-start data for mathematical reasoning following the procedure in Section~\ref{sec:data_construction}, producing approximately $43$K query-response pairs.

\paragraph{Training} We perform format SFT on the Qwen3-4B model for $400$ steps using a learning rate of $2 \times 10^{-6}$ and a batch size of $128$. Then, we apply reinforcement learning, where we sample asynchronous thinking trajectories with an agent pool capacity of $c=4$. We use a learning rate of $10^{-6}$ and a batch of $128$, and sample $8$ trajectories for each query to compute group-relative advantages.
We set the maximum response length of \our{} \worker{}s to $512$ tokens when answering each sub-query, and reset it to $512$ each time \our{} \org{} finishes a \texttt{Join} action.

\paragraph{Baselines} 
We compare \our{} against sequential and parallel thinking baselines, which train the model with DeepSeek-R1-style reinforcement learning. Specifically, sequential thinking performs reasoning within a single sequential thinking trace; parallel thinking first generates four thinking trajectories independently and then performs majority voting to determine the final answer. The suffix `L1K' indicates that the maximum response length allowed during reinforcement learning is $1$K tokens, and the same convention applies to `L2K'.

\begin{table*}[t]
\centering
\scalebox{1.0}{
\begin{tabular}{lcc|cc}
\toprule
\multirow{2}{*}{\bf Methods} & \multicolumn{2}{c|}{AIME-24} & \multicolumn{2}{c}{AMC-23} \\
& Accuracy~($\uparrow$) & Latency~($\downarrow$) & Accuracy~($\uparrow$) & Latency~($\downarrow$) \\
\midrule
Sequential-Thinking-L1K & 24.7 & 1022.6 & 59.5  & 990.0 \\
Sequential-Thinking-L2K & 35.3 & 2048.0 & 67.0 & 2001.1 \\
Parallel-Thinking-L1K & 24.7  & 1024.2 & 61.9 & 1029.5 \\
Parallel-Thinking-L2K & \bf 38.7  & 2048.0 & 72.8  & 2031.4 \\
\our{}      & \bf 38.7  & 1468.0 & \bf 73.3  & 1459.5 \\
\bottomrule
\end{tabular}
}
\caption{Math reasoning evaluation results on AIME-24 and AMC-23. We report the accuracy and critical-path latency results for each benchmark. Results are averaged across $5$ random seeds.}
\label{tab:math_main}
\end{table*}

\paragraph{Results}
Table~\ref{tab:math_main} presents the evaluation results on AIME-24 and AMC-23. \our{} obtains the best overall performance with substantially lower critical-path latency than the baseline methods. In particular, each \our{} \worker{} is restricted to a response length of $512$ tokens for each sub-query, which is substantially shorter than the $2$K thinking traces used in the baselines, but overall performance remains competitive. The results indicate that short, organized thinking fragments can collectively achieve high problem-solving quality under the learned organization policy.
This highlights the importance of the critical-path latency: the communication between \worker{}s and the \org{} introduces non-trivial overhead (e.g., 512 \worker{}-wise limits vs. roughly 1.5K overall latency), suggesting that future research should explicitly consider such overhead when evaluating non-sequential thinking methods.

\begin{table*}[t]
\centering
\scalebox{1.0}{
\begin{tabular}{lccc|ccc}
\toprule
\multirow{2}{*}{} & \multicolumn{3}{c|}{MCD} & \multicolumn{3}{c}{AMC-23} \\
& Accuracy~($\uparrow$) & $\eta$~($\uparrow$) & Latency~($\downarrow$) & Accuracy~($\uparrow$) & $\eta$~($\uparrow$) & Latency~($\downarrow$) \\
\midrule
\our{}      & \bf 89.0 & 64.7 & 4525.4  & \bf 73.3 & 44.8 & 1459.5 \\
~~$-$$R_\eta$ Reward & 85.3 & 61.3 & 6250.6 & 71.3 & 26.1 & 1933.3 \\
~~$-$Format SFT & 64.8 & 50.0 & 3433.6 & 54.9 & 25.0 & 1007.7 \\ 
~~$-$RL &  0.0 & 49.2 & 1987.6 & 3.6 & 33.3 & 1396.9 \\
\bottomrule
\end{tabular}
}
\caption{Ablation study results by removing key components of \our{}. Results are averaged across $5$ random seeds.}
\label{tab:ablation}
\end{table*}

\begin{table*}[t]
\centering
\scalebox{1.0}{
\begin{tabular}{lcc}
\toprule
\bf Methods & Accuracy~($\uparrow$) & Latency~($\downarrow$) \\
\midrule
Sequential-Thinking      & 65.7  & 2055.5 \\
Parallel-Thinking & 84.2 & 3694.7 \\
\our{} & \bf 89.4 & 2853.0 \\
\bottomrule
\end{tabular}
}
\caption{Evaluation on an out-of-domain task, i.e., training on countdown data but evaluating on the $4\times4$ Sudoku task. \our{} generalizes its asynchronous thinking capability to Sudoku, achieving higher accuracy.}
\label{tab:sudoku}
\end{table*}

\subsection{Generalization to Unseen Task}
We study whether \our{} learns the generalized organization policy beyond the domain of training data. Specifically, we directly evaluate the \our{} and the baseline models, which are previously trained on the multi-solution countdown task, on the $4\times4$ Sudoku task. We sample $400$ $4\times4$ Sudoku examples from the Enigmata dataset as a test set \cite{chen2025enigmata}. The evaluation results are shown in Table~\ref{tab:sudoku}. Remarkably, \our{} obtains superior Sudoku performance while maintaining a much lower latency compared to parallel thinking. The results demonstrate that \our{} can learn general asynchronous thinking capabilities beyond the trained tasks.

Beyond evaluating the \our{} model on Sudoku, we further assess its generalization on two out-of-domain queries from computer science and biology. The corresponding thinking trajectories are presented in Appendix~\ref{app:cases}.

\subsection{Ablation Studies}
\paragraph{Effect of Format Fine-Tuning}
We examine the effect of cold-start format fine-tuning in \our{}. For comparison, we skip the format fine-tuning stage and directly train the model with the same reinforcement learning recipes as \our{} on the multi-solution countdown and math reasoning tasks, respectively. As shown in Table~\ref{tab:ablation}, the `$-$Format SFT' variant achieves much lower accuracy than \our{} for both tasks. Interestingly, we observe that the trained models can successfully perform \texttt{Fork}s but consistently remain the thinking concurrency ratio of $\eta = 1/c$, reflecting as $50.0\%$ and $25.0\%$ concurrency ratios in the two tasks, respectively. This behavior indicates that the cold-start format learning stage establishes a good initialization for \our{} to learn to explore and exploit asynchronous thinking structures.

\begin{figure*}[t]
    \centering
    \includegraphics[width=1.0\linewidth]{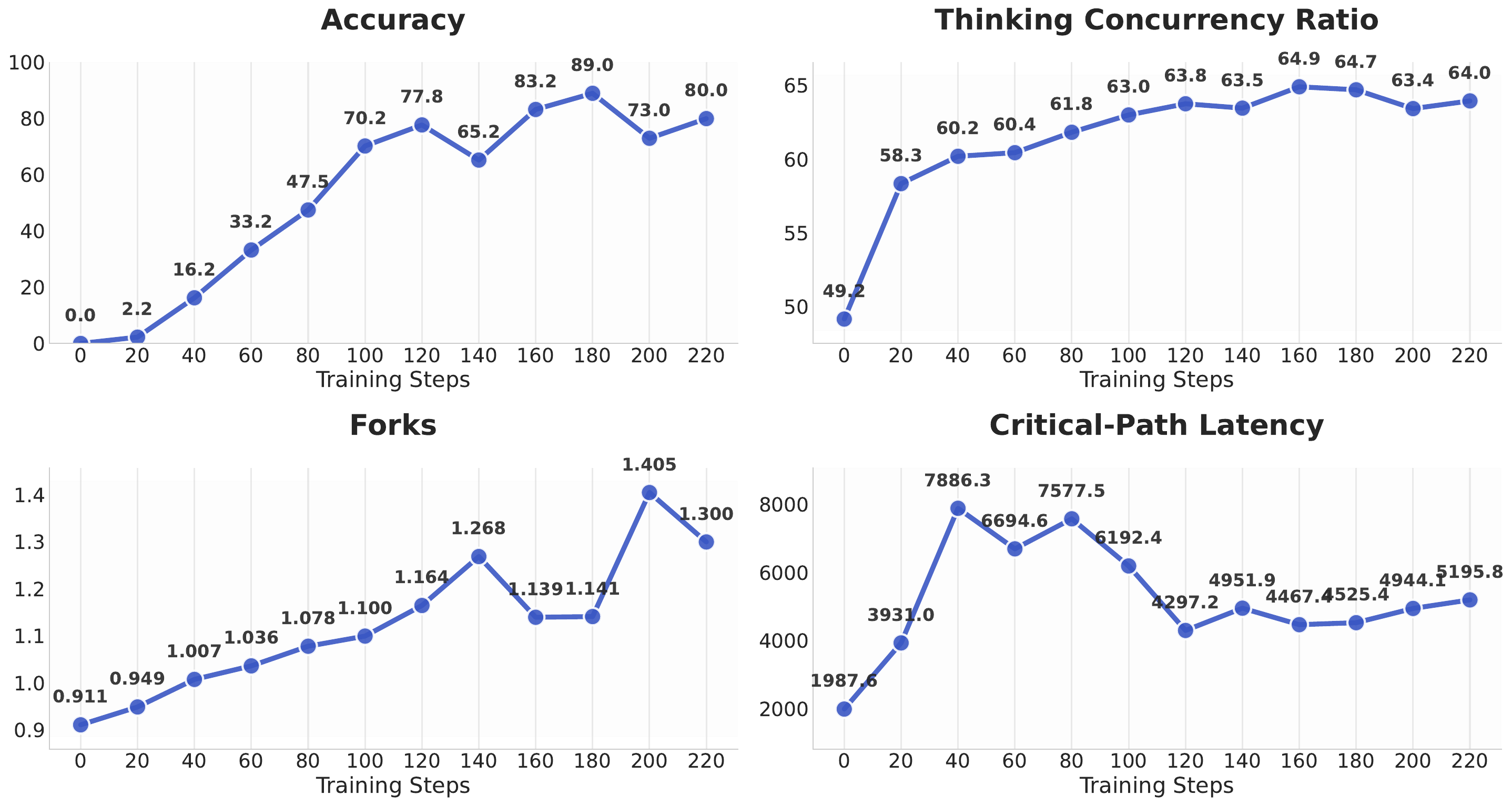}
    \caption{Trajectories of accuracy, thinking concurrency ratio, average number of \texttt{Fork}, and critical-path latency throughout training.}
    \label{fig:eff_rl}
\end{figure*}

\paragraph{Effect of Reinforcement Learning}
Figure~\ref{fig:eff_rl} illustrates the training trajectories of accuracy, thinking concurrency, average number of \texttt{Fork} operations per query, and critical-path latency during \our{} reinforcement learning. Accuracy starts at zero and increases steadily, reaching $89.0\%$ after $180$ training steps. During training, the critical-path latency increases rapidly and reaches a peak at the enforced upper limit of $8$K, reflecting that the model initially tends to think longer. As training progresses, the model learns to produce more concurrently-executable parts, leading to a steady increase in thinking concurrency and a decrease in latency. Meanwhile, the number of \texttt{Fork} operations per query also increases during training, indicating that the model progressively learns to distribute the thinking across the workers rather than a single thinking chain. These results highlight the benefits of modeling the thinking structure as model-predictable actions, allowing the model to adapt, explore, and optimize its thinking behavior according to task demands.

\paragraph{Reward Modeling}
To study the effects of the reward designs of \our{}, we conduct reinforcement learning using different rewards, and keep the other setups fixed. As shown in Table~\ref{tab:ablation}, `$-$$R_\text{U}$ Reward' stands for removing the thinking concurrency reward. Removing this reward reduces accuracy and leads to higher latency on both tasks, confirming its effect in encouraging parallelism among \worker{}s and the \org{}. In another experiment, we design an additional leverage reward, which rewards the \org{} to do \texttt{Fork} operations after a previous \texttt{Join}, building upon prior merged thinking outcomes. However, the reward introduces instability during training, sometimes resulting in a collapse to a fixed mode that exhibits near-sequential thinking with interleaved \texttt{Fork}s and \texttt{Join}s.

\paragraph{Accuracy-Latency Frontier}

\begin{figure*}[t]
    \centering
    \includegraphics[width=0.55\linewidth]{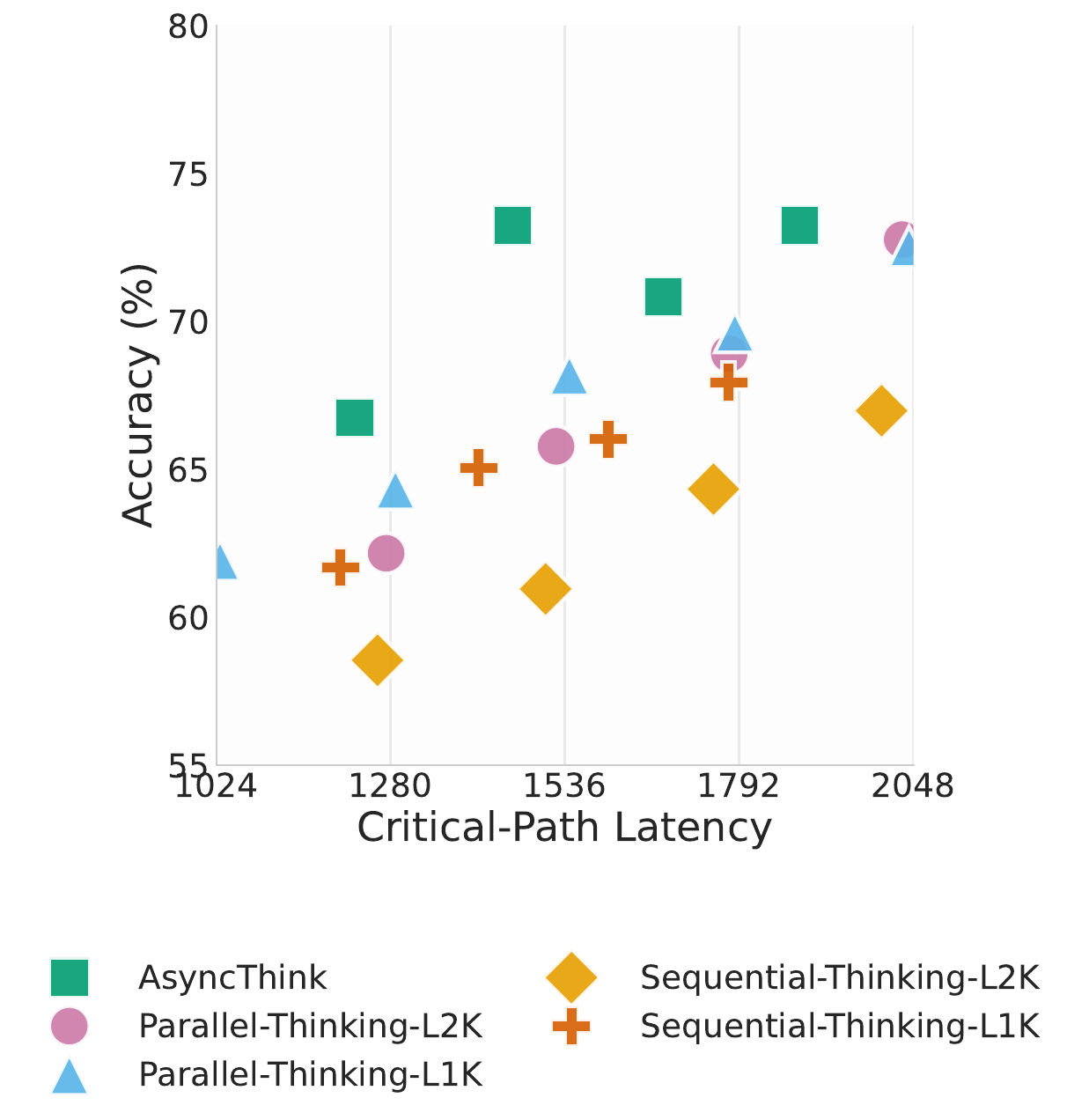}
    \caption{Accuracy-latency frontier of \our{} and baseline methods. The data points are collected by inference under various configurations of maximum response length.}
    \label{fig:acc-latency-frontier}
\end{figure*}

Figure~\ref{fig:acc-latency-frontier} illustrates the accuracy-latency trade-off for \our{} and baseline methods. Each marker type represents a distinct method. We set various maximum response length limits for baseline methods to obtain the accuracy scores and their corresponding latencies. For \our{}, we keep the maximum response length of \org{} fixed and set various maximum response lengths for \worker{}s to obtain accuracy scores with different overall latencies. Figure~\ref{fig:acc-latency-frontier} shows that \our{} achieves a superior accuracy-latency frontier. In particular, \our{} reduces inference latency by $28\%$ compared to parallel thinking while still achieving higher math reasoning accuracy. These results demonstrate that \our{} can not only successfully perform asynchronous thinking but also yields a more favorable accuracy-latency frontier.

\begin{figure}
    \centering
    \includegraphics[width=\linewidth]{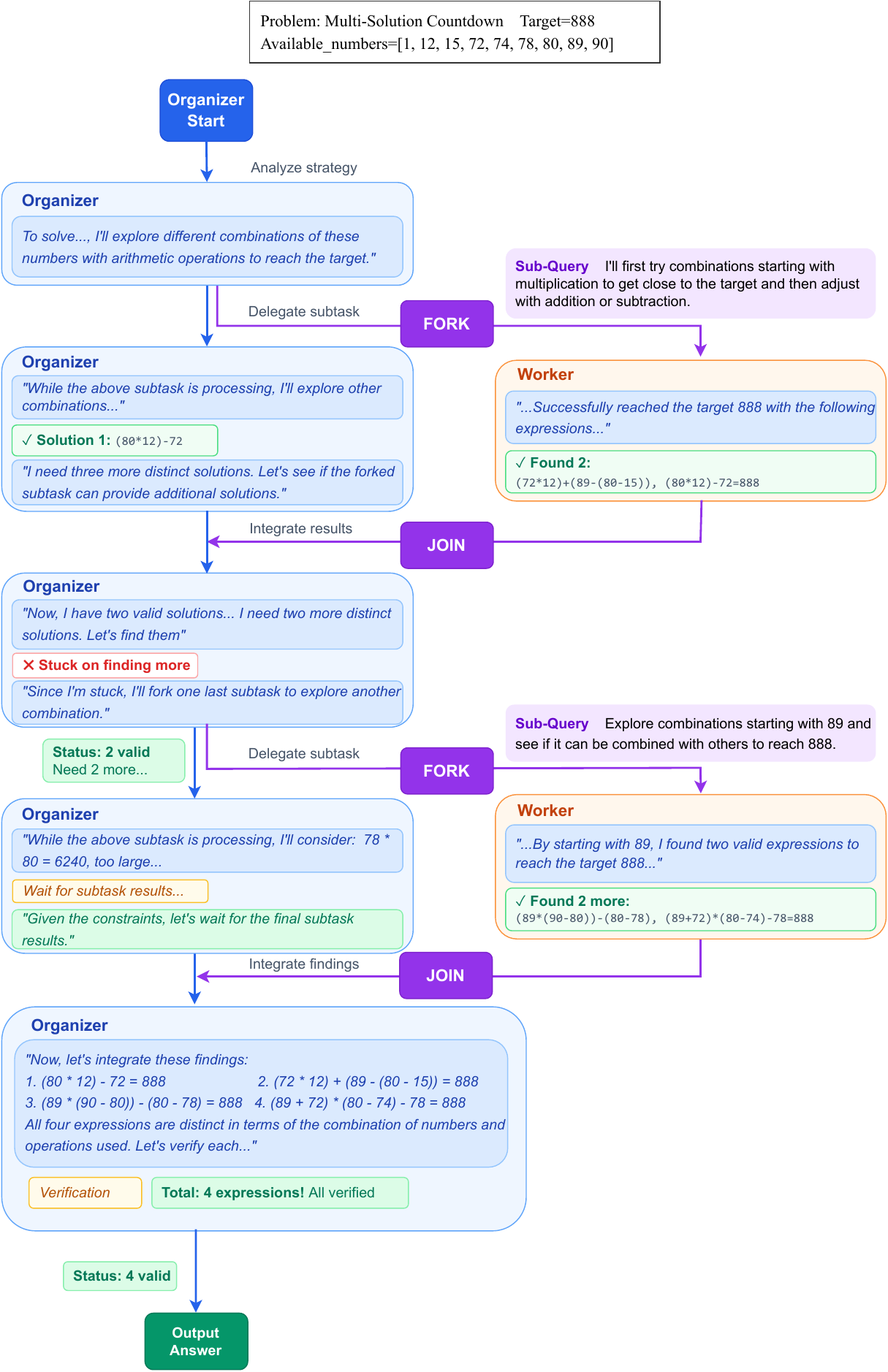}
    \caption{An \our{} thinking trajectory on the multi-solution countdown task with an agent pool capacity of $c=2$. The \org{} has learned to actively perform \texttt{Fork} and \texttt{Join} operations without external intervention. After the first \texttt{Join}, it checks the remaining gaps and launches a new sub-query as needed. The process finally yields four distinct valid expressions.}
    \label{fig:casecnt}
\end{figure}

\subsection{Case Studies}

To further illustrate how the learned \our{} models organize thinking, we present two representative examples in Figure~\ref{fig:casecnt} and Figure~\ref{fig:casemath}. These examples show how \our{} structures the thinking processes with novel asynchronous patterns beyond sequential thinking.

In the multi-solution countdown task (Figure~\ref{fig:casecnt}), our \our{} model learns to perform multistage divide-and-merge reasoning, where the \org{} adaptively invokes a \worker{} to search for valid solutions of the input question. After analyzing the task, the \org{} assigns a sub-query to a \worker{} to first explore \textit{multiplication-based} combinations toward the target number.
Simultaneously, the \org{} continues to explore other combinations. Once the \worker{} completes, the results are merged into the \org{}, and the \org{} assigns new sub-queries when necessary. This iterative \texttt{Fork}-\texttt{Join} process enables efficient exploration and produces four valid expressions that reach the target of $888$. In contrast, sequential thinking failed to find all four valid solutions in the case, highlighting the ability of \our{} to organize efficient thinking.

In the case of math reasoning (Figure~\ref{fig:casemath}), \our{} shows an interesting structure. With an agent pool of $c=4$, the \org{} assigns sub-queries to three \worker{}s, each exploring a distinct direction.  
For example, \worker{}~3 is instructed to ``assume the regular tetrahedron edge length can be set for simplicity.''  
These \worker{}s run concurrently and their results are \texttt{Join}-ed back once the \org{} verifies consistency across conclusions.

Across the case studies, \our{} distributes the thinking into disciplined \texttt{Fork}-\texttt{Join} exploration and merges them into a coherent answer. It achieves broader solution coverage and higher reliability with lower latency than sequential thinking. Moreover, it showcases organized reasoning patterns beyond reflection or self-correction, demonstrating an ability to organize multiple thinking paths and leverage intermediate results to conduct further thinking.

\section{Related Work}

\paragraph{Chain-of-Thought Reasoning}
Chain-of-thought (CoT) reasoning \cite{cot} refers to generating an explicit sequence of natural-language thinking steps that culminates in a final answer. To further strengthen the reasoning capability of large language models (LLMs), reinforcement learning with verifiable rewards (RLVR) has emerged as a key post-training paradigm. This framework enables models to receive learning signals from automatically verifiable outcomes rather than subjective human feedback. Recent RLVR-based models, such as DeepSeek-R1 \cite{r1}, demonstrate that scalable verifiable supervision can substantially improve reasoning capabilities.
Despite these advances, the design of verifiable reward functions remains an open problem \cite{zhang2025survey}. Recent work has proposed model-based verifiers \cite{ma2025general,liu2025compassverifier}, reasoning reward models \cite{rrm,j1}, generator-reward co-evolving \cite{rltango,zhang2025critique}. In addition, current research also explores a range of policy optimization algorithms, including critic-based algorithms \cite{ppo,vapo}, critic-free algorithms \cite{grpo,rloo}, off-policy methods \cite{cohen2025soft,roux2025tapered}, etc.

\paragraph{Parallel Thinking}
Although longer reasoning traces can improve model performance, many studies show that generating multiple reasoning paths independently and aggregating their results also improves performance \cite{yao2023tree,huang2025efficient,zhao2025majority,brown2024large,rrm,r1,wen2025parathinker}. To overcome the inherent sequential nature of LLM decoding, recent work has explored parallel decoding algorithms and model architectures that allow simultaneous generation across branches or tokens \cite{jinlearning,rodionov2025hogwild,nie2025large,hsu2025group,yang2025multiverse}. 
However, prior approaches often depend on handcrafted coordination rules or imitation from curated SFT data, and do not support reinforcement learning to further acquire reasoning capabilities.
Beyond that, APR~\cite{apr} shows that LLMs can achieve parallel reasoning by imitating the execution traces of depth first search (DFS) and breadth first search (BFS) algorithms. However, this approach relies on predefined algorithmic traces, which limits its applicability to open-ended reasoning tasks where no universal DFS or BFS procedure can be specified. 
More recently, concurrent work Parallel-R1~\cite{parallel-r1} demonstrated that LLMs can learn to actively start sampling parallel thinking traces in the middle and aggregate their outcomes through model-driven decisions rather than rule-based schemes such as majority voting. Differently, we propose \our{} capable of learning to organize thinking processes, assigning diverse sub-queries and achieving lower latency compared to parallel thinking.

\paragraph{Multi-Agent Systems}
LLM-based multi-agent systems employ multiple agents to collaboratively solve complex tasks via specialized roles and iterative dialogue~\cite{Li2024ASO,Li2023TrainerAgentCA,Barbarroxa2024BenchmarkingLL}.
A wide range of multi-agent systems operate via predefined conversational roles~\cite{metagpt,gamegpt}.
Such designs facilitate structured task execution through cooperative dialogues~\cite{autogen,Li2023CAMELCA} or competitive debates~\cite{debate,debate2}.
Recent advances demonstrate that multi-agent LLMs can achieve more adaptive and efficient collaboration through dynamic belief modeling~\cite{Zhang2023ProAgentBP}, role exchange~\cite{ma2024coevolving}, and agent co-evolution~\cite{Hu2024SelfEvolvingMC,Belle2025AgentsOC}.
To further enhance stability and efficiency, \citet{Zahedifar2025LLMAgentControllerAU} integrates a central controller agent for high-level planning, and \citet{Suzgun2024MetaPromptingEL} guides the meta agent in assuming distinct sub-roles through meta-prompting.
A concurrent work, Puppeteer~\cite{Dang2025MultiAgentCV}, further explores RL-based dynamic orchestration of LLM agents, providing additional evidence that collective intelligence can enhance model capabilities through specialized roles and cross-verification.

\section{Conclusion and Future Work}

\paragraph{Scaling Agentic Organization with Massive Agents}
This has two primary dimensions.
First is scaling the quantity of workers. Future work should explore the scaling laws of asynchronous thinking: how accuracy–latency trade-offs evolve as the agent pool capacity grows from a few to hundreds or even thousands.
Second is scaling the diversity of agents. We can move beyond a homogeneous pool to a massive organization of heterogeneous expert workers. These agents could be fine-tuned for specific domains (e.g., math, coding, data analysis) and, crucially, be equipped with different external tools (such as code interpreters, database query engines, or web search APIs). This introduces a more complex and powerful learning problem for the organizer.

\paragraph{Recursive Agentic Organization}
In this paradigm, any worker could dynamically be promoted to a sub-organizer, gaining the ability to Fork its own team of sub-workers. This would enable a flexible, hierarchical structure, naturally suited for deeply nested and complex problems that require multi-level decomposition. For instance, a top-level organizer might delegate a broad query like ``\textit{Solve the * problem},'' only for the assigned worker to act as a sub-organizer, forking three new sub-workers to independently test different lemmas in parallel.

\paragraph{Human-AI Agentic Organization}
A key frontier is creating a Human-AI collaborative framework by integrating humans directly into the agentic organization. This could involve a ``Human-as-Organizer'' using the Fork protocol to dispatch tasks to AI workers, or a ``Human-as-Worker'' where the AI Forks tasks requiring human judgment (e.g., <FORK-human>Verify this conclusion</FORK-human>). Additionally, collaborative planning would allow humans and AI to co-design the asynchronous strategy before execution. This path moves beyond pure AI autonomy to enable a powerful, hybridized intelligence.

\newpage
\bibliography{example_paper}
\bibliographystyle{plainnat}

\appendix

\newpage
\section{\our{} Thinking Cases}
\label{app:cases}

\begin{figure}
    \centering
    \includegraphics[width=\linewidth]{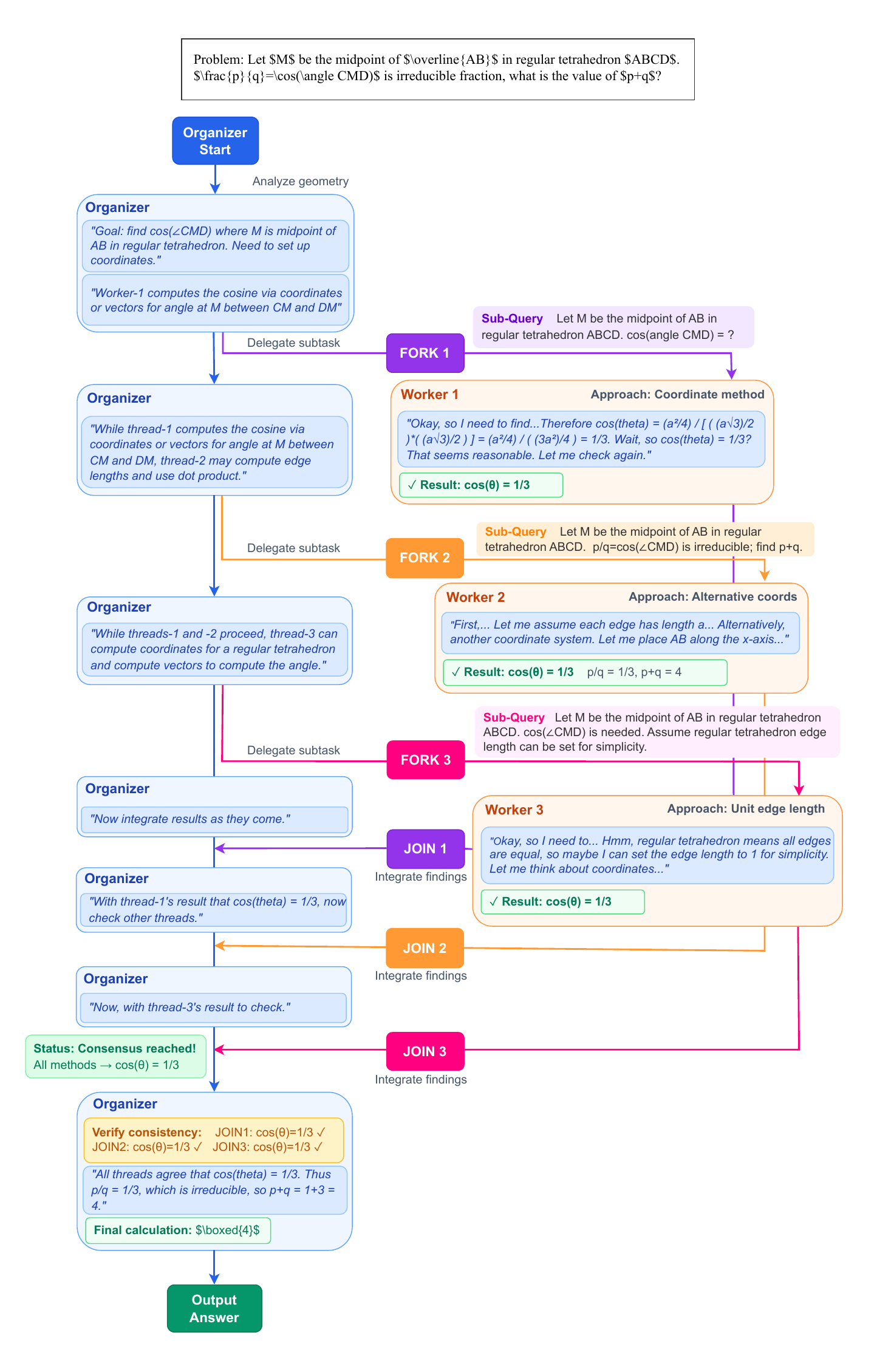}
    \caption{An \our{} thinking trajectory on mathematical reasoning with an agent pool capacity $c=4$. The \org{} spawns three \worker{}s, each using a different geometric formulation. These \worker{}s run in parallel and derive consistent results ($\cos \theta = 1/3$). They then rejoin for verification, converging to the final answer.}
    \label{fig:casemath}
\end{figure}

\begin{figure}
    \centering
    \includegraphics[width=\linewidth]{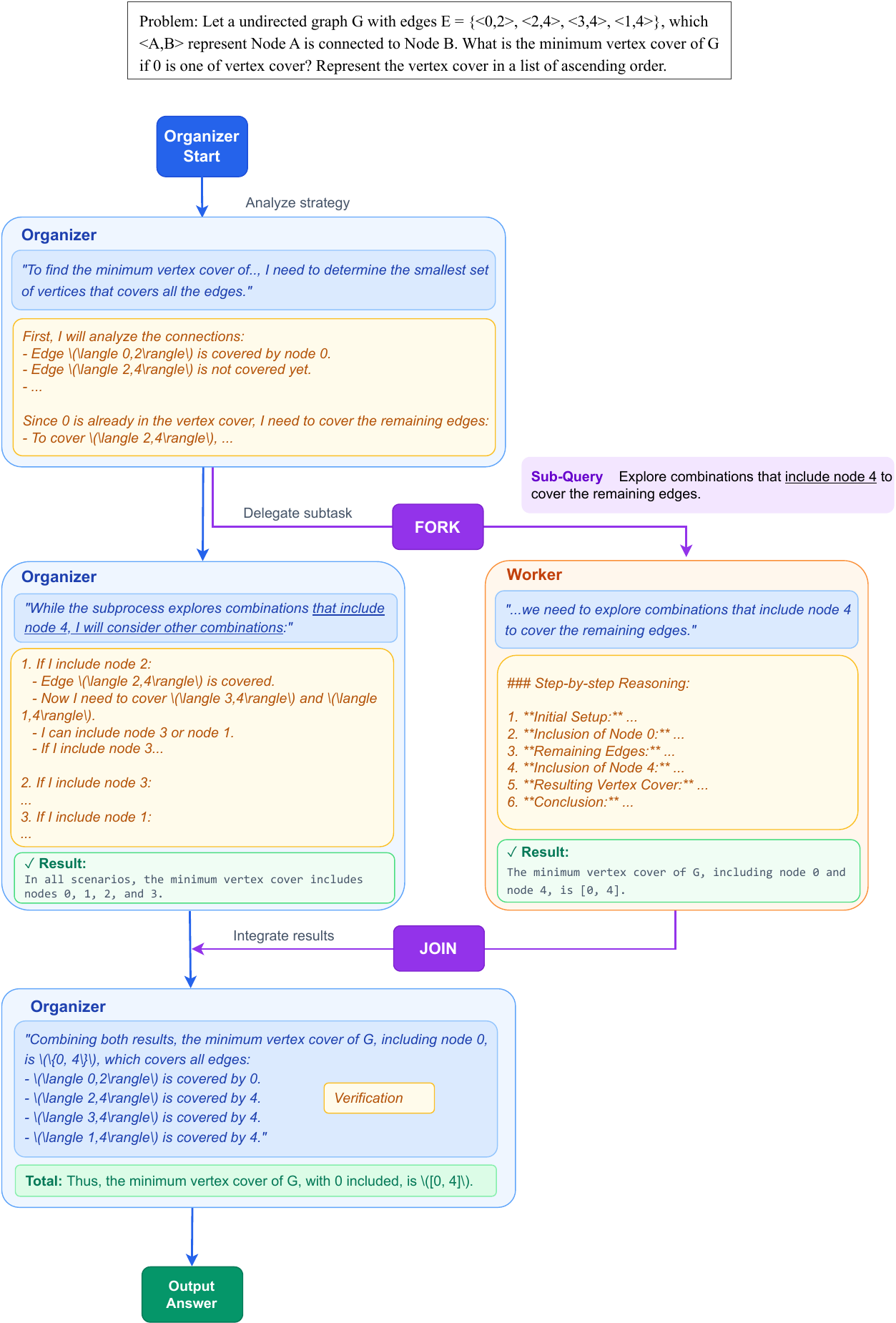}
    \caption{An example of \our{} on a minimum vertex cover problem from MMLU-Pro (with agent pool capacity $c=2$). The \org{} generalizes to this unseen graph theory task, performing \texttt{Fork} and \texttt{Join} operations without prior training on such problems. While a \worker{} explores combinations with node 4, the \org{} concurrently analyzes alternatives, ultimately determining the minimum vertex cover [0, 4].}
    \label{fig:casecs}
\end{figure}

\begin{figure}
    \centering
    \includegraphics[width=\linewidth]{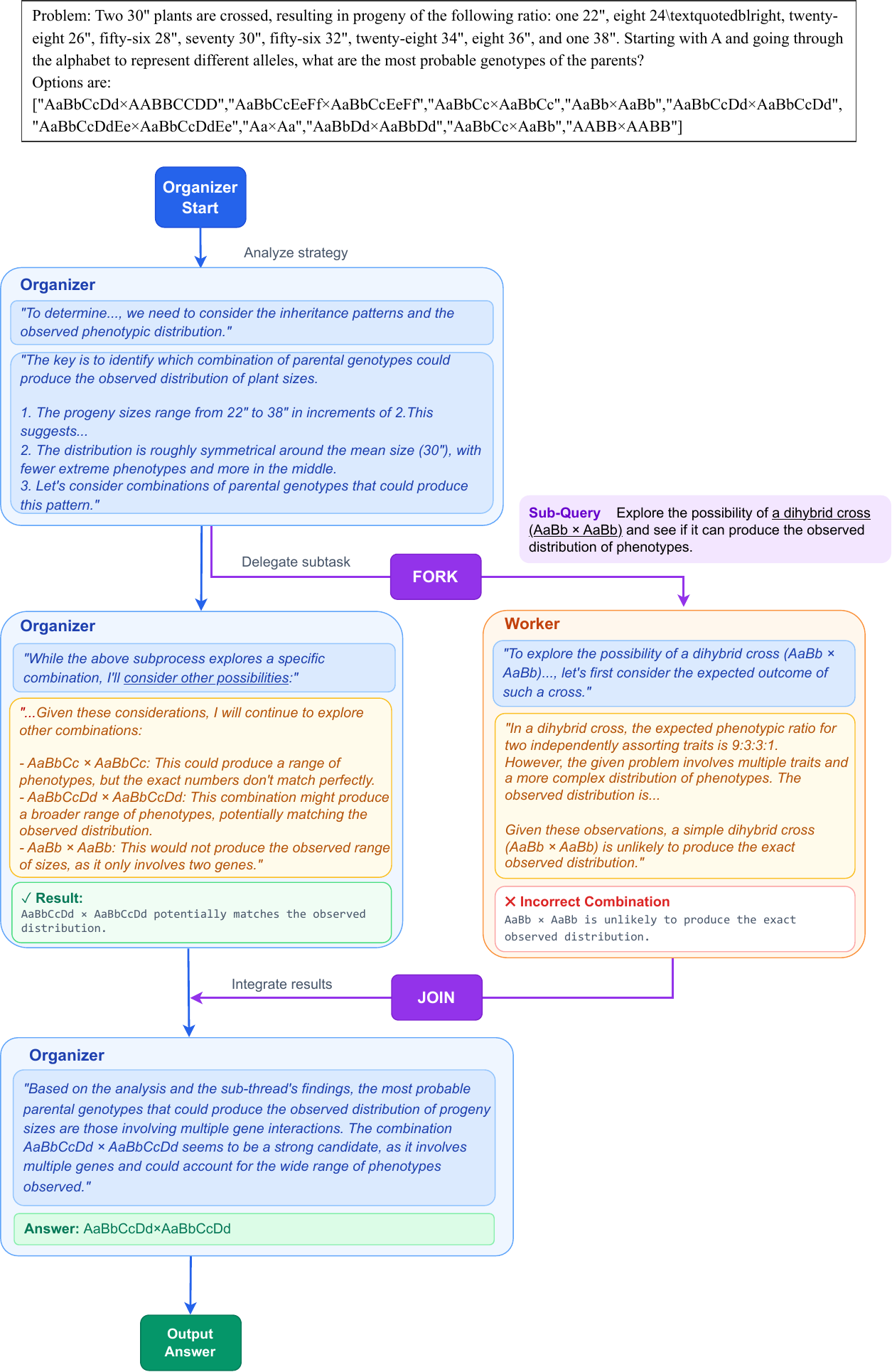}
    \caption{An example of \our{} on a genetics cross problem from MMLU-Pro (with agent pool capacity $c=2$). The \org{} generalizes to this biology domain, performing \texttt{Fork} and \texttt{Join} operations without domain-specific training. While a \worker{} explores the dihybrid cross (AaBb × AaBb), the \org{} concurrently evaluates multi-gene combinations, ultimately identifying AaBbCcDd × AaBbCcDd as the most probable parental genotypes.}
    \label{fig:casebio}
\end{figure}

We provide several \our{} thinking trajectories. Figure~\ref{fig:casemath} presents an \our{} thinking trajectory on math reasoning. Figure~\ref{fig:casecs} and Figure~\ref{fig:casebio} present the thinking trajectories of the \our{} model, trained on the multi-solution countdown data, demonstrating its ability to generalize its asynchronous thinking capabilities to out-of-domain tasks.

\section{Prompt Templates}

We provide the prompt templates for organizer and worker of \our{} as follows. `\{\dots\}' represents a variable placeholder in the prompt template. For example, `\{Capacity $-1$\} means this placeholder should be replaced with the agent pool capacity minus one, according to the experimental setup.

\begin{tcolorbox}[
  title=Prompt Template for Organizer ,
  colback=blue!5,
  colframe=blue!50!white,
  fonttitle=\bfseries,
  coltitle=white,
  colbacktitle=gray!70!blue,
  label={box:organizer-prompt},
  enhanced,
  breakable,
  boxrule=0.5pt,
  arc=2mm,
  listing only,
  listing options={
    basicstyle=\ttfamily\small,
    breaklines=true,  
    postbreak=\mbox{\textcolor{red}{$\hookrightarrow$}\space}, 
    escapeinside={(*@}{@*)}, 
  }
]
This is main process. \\

\{Task-specific instruction here\}  \\

Use <FORKi>subtask description</FORKi> to delegate work and <JOINi> to wait for results. Integrate these results and provide final answer with <ANSWER>your final answer</ANSWER>. Notice that you can have at most \{Capacity $- 1$\} subtasks running concurrently; any additional ones must wait until earlier ones finish. \\

\{Query\}
\end{tcolorbox}

\begin{tcolorbox}[
  title=Prompt Template for Worker ,
  colback=blue!5,
  colframe=blue!50!white,
  fonttitle=\bfseries,
  coltitle=white,
  colbacktitle=gray!70!blue,
  label={box:worker-prompt},
  enhanced,
  breakable,
  boxrule=0.5pt,
  arc=2mm,
  listing only,
  listing options={
    basicstyle=\ttfamily\small,
    breaklines=true,  
    postbreak=\mbox{\textcolor{red}{$\hookrightarrow$}\space}, 
    escapeinside={(*@}{@*)},  
  }
]

This is a subprocess. \\

\{Task-specific instruction here\} \\

Subtask: \{Sub-query assigned by the organizer\} \\

Complete the subtask and provide results in: \\
<RETURN> \\
(Your short summary and valid expressions found) \\
</RETURN> \\
\end{tcolorbox}

\end{document}